\documentclass[11pt]{article}

\usepackage[final]{acl}

\usepackage{times}
\usepackage{latexsym}

\usepackage[T1]{fontenc}

\usepackage[utf8]{inputenc}

\usepackage{microtype}

\usepackage{inconsolata}

\usepackage{graphicx}

\usepackage{amssymb}
\usepackage{amsmath}
\usepackage{booktabs}
\usepackage{multirow}
\usepackage{subscript}
\usepackage{xcolor}
\usepackage{xstring}
\usepackage{pifont}
\usepackage{arydshln}
\usepackage{hyperref}
\newcommand{\cmark}{\ding{51}}
\newcommand{\xmark}{\ding{55}}

\title{Evaluation Pitfalls and Challenges in Multimedia Event Extraction}

\author{
 \textbf{Philipp Seeberger}, 
 \textbf{Steffen Freisinger}, 
 \textbf{Tobias Bocklet},
 \textbf{Korbinian Riedhammer}
\\
 Technische Hochschule Nürnberg Georg Simon Ohm \\
  \small$\texttt{\{philipp.seeberger,steffen.freisinger,tobias.bocklet,korbinian.riedhammer\}@th-nuernberg.de}$
}

\begin{document}
\maketitle
\begin{abstract}
Multimedia event extraction aims to jointly identify events and their arguments across multiple modalities, such as text and images, to support more comprehensive event understanding.
While recent work reports steady and substantial progress, the reliability and comparability of these results critically depend on consistent and rigorous evaluation.
In this work, we present the first systematic analysis of evaluation pitfalls in multimedia event extraction and identify three major sources of issues: inconsistent data processing, inconsistent task assumptions, and overly relaxed evaluation settings.
We demonstrate, through a series of controlled experiments under a strict evaluation framework, that minor evaluation choices can cause large performance variations and lead to overestimation of a model’s ability to ground real-world events across modalities.
Our findings highlight the need for comparable evaluation standards and encourage a shift toward more rigorous evaluation in multimedia event extraction.\footnote{\url{https://github.com/seebergerph/StrictEval}}
\end{abstract}

\section{Introduction}\label{sec:introduction}

Event extraction (EE) is a fundamental task in natural language processing and information extraction, aiming to identify, structure, and organize event-related knowledge from documents \cite{ahn2006}.
While the majority of existing EE research has focused on texts \cite{omnievent2023,textee2024}, recent work has increasingly explored the integration of additional modalities \cite{umie2024,muie2024}.
This shift is motivated by the growing prevalence of multimodal content in contemporary news media and online platforms \cite{wase2020}, where images, videos, and audio provide complementary information that can support more accurate and comprehensive event understanding.

Prior research has investigated EE or closely related tasks within individual modalities \cite{imsitu2016,dygie2019,vidsitu2021,speechEE2024}, including text, images, video, and audio, or has leveraged cross-modal cues to address specific challenges such as ambiguity \cite{vad2017,imageEnhanced2020}.
However, evaluation mostly remains restricted to a single target modality.
Multimedia event extraction (MEE) has recently attracted attention and adopts a holistic view by jointly extracting and evaluating events across multiple modalities, typically combining textual and visual inputs \cite{wase2020,vm2e2_2021,multivent2024}.
Despite this progress, existing MEE benchmarks  remain limited and evaluation challenging, most notably due to annotation scarcity, lack of train splits, and increased evaluation complexity inherent to multimodal settings \cite{wase2020,multivent2024}.

Prior studies have demonstrated that even traditional textual EE suffers from substantial and often overlooked evaluation challenges \cite{revisitingEE2021,pitfalls2023,textee2024}, making it prone to hidden pitfalls.
These include discrepancies in data and task assumptions as well as metric design choices that can distort model comparisons and fail to reflect real-world performance \cite{textee2024}.
Crucially, extending textual EE to the multimodal setting not only inherits existing evaluation issues but also introduces additional pitfalls.
These arise from factors such as data scarcity, heterogeneous modalities, and multi-stage pipelines commonly employed in MEE \cite{wase2020,unicl2022,camel2023,xmtl2025}.
As a consequence, inconsistent and under-specified evaluation settings can easily emerge, posing a potential obstacle to reliably assessing progress in MEE research.

Motivated by concerns about the reliability and comparability of current evaluations, this work systematically investigates hidden pitfalls and challenges in MEE evaluation, with the goal of raising awareness and encouraging a shift toward more rigorous evaluation practices.
Through an in-depth analysis of the widely used M2E2 benchmark, we first identify three major categories with several issues: inconsistent data processing, inconsistent task assumptions, and relaxed evaluation settings.
Building on this analysis, we introduce a more rigorous evaluation framework, \textsc{StrictEval}, and use it to examine how hidden pitfalls influence reported performance. 
Finally, we show that minor experimental design choices can substantially affect evaluation outcomes.

In summary, our contributions are twofold: 
(1) We conduct a systematic analysis of evaluation pitfalls and challenges in MEE and propose a more rigorous evaluation framework (\textsc{StrictEval}).
(2) We systematically quantify how hidden evaluation pitfalls affect reported performance and reevaluate recent MEE approaches to highlight limitations.
\section{Background and Related Work}\label{sec:related}

\subsection{Background}\label{subsec:background}
\begin{figure*}
    \centering
    \includegraphics[width=\linewidth]{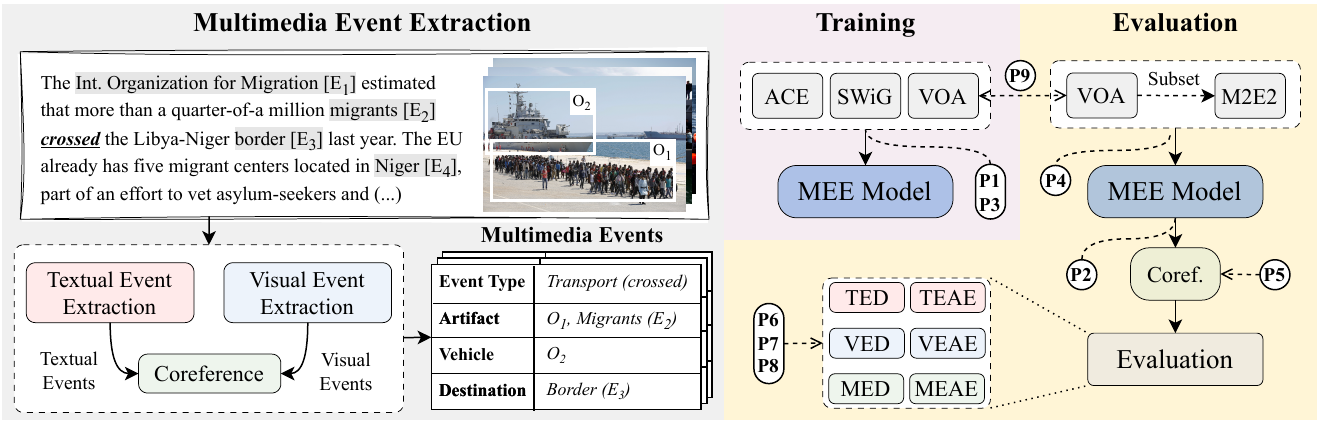}
    \caption{Overview of the M2E2 multimedia event extraction pipeline. The example illustrates a \textit{Transport} event grounded in text and image. \textsc{P} markers indicate stages at which pitfalls occur. TED, TEAE, VED, VEAE, MED, and MEAE denote the textual, visual, and multimedia event detection and argument extraction subtasks, respectively.}
    \label{fig:framework}
\end{figure*}

Textual EE is commonly formulated as a two-stage pipeline \cite{ahn2006} consisting of event detection (ED) and event argument extraction (EAE). 
Event detection aims to identify event mentions, typically grounded to trigger spans, and classify them into predefined event types. Event argument extraction focuses on identifying argument spans and assigning them semantic roles conditioned on the detected event mentions.
Analogously, visual EE decomposes the task into detecting events grounded in images and linking their associated semantic roles to visual regions, such as objects \cite{swig2020}.
Building on these two research directions, MEE integrates textual and visual information to jointly extract events and their arguments across modalities \cite{wase2020,vm2e2_2021}.
This multimodal integration introduces the additional subtask of cross-modal event coreference resolution, with the aim to unify event mentions from different modalities that refer to the same real-world event into a coherent multimedia event representation (see \autoref{fig:framework}).

\subsection{Related Work}\label{subsec:related_work}

\paragraph{Multimedia Event Extraction Benchmarks} While most EE benchmarks focus exclusively on text \cite{ace2006,ere2015,maven2020,textee2024}, early multimodal extensions augment textual datasets with images, but evaluation remains limited to textual events \cite{vad2017,imageEnhanced2020}.
To overcome unimodal limitations, \citet{wase2020}
introduce the first MEE benchmark, M2E2, which evaluates event and argument extraction for both texts and images.
In addition, M2E2 includes cross-modal event coreferences, analogous to cross-document coreferences in text \cite{crossDocCoref2024}.
Subsequent work extend to images and videos, such as VM2E2 \cite{vm2e2_2021}, CMMEvent \cite{vegsrf2025}, TVEE \cite{tvee2023}, and MultiVENT-G \cite{multivent2024}.
More recently, \citet{muie2024} propose a comprehensive benchmark covering textual, visual, and audio inputs by integrating datasets such as M2E2 and ACE with recorded speech.
However, only M2E2 and MultiVENT-G publicly release complete data while other benchmarks are still closed-source \cite{tvee2023,vegsrf2025} or lack critical annotations \cite{vm2e2_2021,muie2024}.
Moreover, complex annotation formats, abundance of train splits, and missing evaluation scripts further hinder reliable benchmarking.

\paragraph{Multimedia Event Extraction}
Early approaches focus on cross-modal correlations and align visual and textual representations using large-scale unlabeled news corpora (e.g., VOA) \cite{vm2e2_2021,unicl2022,mgim2024}, often in combination with contrastive learning objectives \cite{wase2020,clipevent2022,tsee2023}. Subsequent studies explore complementary directions, including augmenting training data with synthetically generated image–text pairs \cite{camel2023}, designing sophisticated multi-grained fusion mechanisms \cite{mgfsg2025,afssag2025}, or leveraging multi-task learning with pseudo labeling strategies \cite{xmtl2025}.
Other works narrow their focus to specific subtasks, such as ED \cite{mqa2023} or EAE \cite{mmutf2024}.
With recent advances in multimodal large language models (MLLMs), several instruction-following approaches have been proposed to enable more universal information extraction \cite{umie2024,muie2024,clorae2025,ssgpf2025,mllm-ke2025}.
However, most of these methods primarily work with given image–text pairs and do not explicitly address broader MEE settings such as cross-modal event coreference resolution \cite{umie2024,clorae2025,ssgpf2025}.
Notably, the majority of existing approaches are evaluated on the M2E2 benchmark, underscoring its role in advancing MEE research.
Despite substantial progress, prior methods adopt diverse task formulations and evaluation protocols, which hinders fair comparison across different modeling approaches.

\paragraph{Evaluation Pitfalls}
Recent studies have highlighted numerous issues in the evaluation of textual EE models, including inconsistent data assumptions, processing steps, output space discrepancies, and relaxed evaluation metrics \cite{revisitingEE2021,pitfalls2023,omnievent2023,textee2024}.
While these works clearly highlight significant differences in textual EE benchmarking, issues in the evaluation of visual and multimedia EE remains relatively underexplored.

\section{Pitfalls and Challenges in Evaluation}
\label{sec:method}

Motivated by evaluation concerns for MEE, we first present our investigation setup (\S\ref{subsec:preliminaries}) and systematic analysis to identify evaluation issues (\S\ref{subsec:systematic_analysis}).
We then provide a detailed analysis of common pitfalls across three major categories: data processing (\S\ref{subsec:inconsistent_data_processing}), task assumptions (\S\ref{subsec:inconsistent_task_assumptions}), and relaxed evaluation settings (\S\ref{subsec:relaxed_evaluation_settings}).
Lastly, building on the insights from this analysis, we introduce \textsc{StrictEval} as rigorous evaluation framework to address hidden pitfalls (\S\ref{subsec:strict_eval}).

\subsection{Preliminaries}\label{subsec:preliminaries}
To examine the sources and impact of evaluation issues in MEE, we adopt the M2E2 benchmark \cite{wase2020}.
Our choice is motivated by two main considerations: (1) M2E2 is publicly available and, to the best of our knowledge, the most widely used benchmark in MEE research. 
(2) As discussed in \S\ref{subsec:related_work}, alternative benchmarks are often incomplete \cite{vm2e2_2021,muie2024} or not accessible \cite{tvee2023,vegsrf2025}.

\paragraph{M2E2 Dataset} In \autoref{fig:framework}, we show the complete task and involved components.
The M2E2 benchmark comprises 6,167 sentences and 1,014 images from 245 multimedia news documents collected from 108k Voice of America (VOA) documents.
Overall, the events cover 8 event types and 15 argument roles, with 1297 textual and 391 visual events.
Thereby, there exist 309 multimedia events which are coreferenced by 192 textual and 203 visual events.
As no training data exists, the benchmark adopts ACE \cite{ace2006} for textual and imSitu \cite{imsitu2016}, with object groundings from SWiG \cite{swig2020}, for visual EE training.
Annotation mappings to the M2E2 schema are provided by the original work \cite{wase2020}.

\paragraph{M2E2 Evaluation} Following \citet{wase2020}, evaluation is conducted separately for textual, visual, and multimedia EE with precision (P), recall (R), and F1 for the subtasks ED and EAE.
In textual EE, an event mention is correct if its type and trigger offsets match the reference, while arguments must additionally match argument offsets and role types.
For visual EE, a visual event mention is correct if its type matches the reference image, and a visual argument if its event type, role label, and bounding box match a reference argument with IoU $>$ 0.5.
Lastly, a multimedia event mention is correct if its event type and trigger offsets (or image) match the reference trigger (or image).
The inherited textual and visual arguments are evaluated using the same criteria as in the textual and visual modality.
However, in our preliminary analysis we observe inconsistencies in the evaluation criteria for multimedia events, which we discuss in detail in \S\ref{subsec:inconsistent_task_assumptions}.

\subsection{Systematic Analysis}
\label{subsec:systematic_analysis}

We collect peer-reviewed MEE studies evaluated on M2E2 published between 2020 and 2025 through keyword and citation-based searches, resulting in 18 articles across multiple venues (e.g., ACL, ACM, AAAI).
Of these, we analyze 15 studies and exclude three \cite{theia2023,muie2024,xing2025} due to reliance on custom train-test splits or newly introduced evaluation metrics, which hinder direct comparison.
The complete set of methods and reported evaluation scores is summarized in \autoref{tab:methods}.
For each study, we review the article, supplementary materials, and, when available, its public codebase, with particular attention on the training and evaluation stages (see \autoref{fig:framework}).
We focus on data processing, experimental setups, and evaluation protocols \cite{pitfalls2023,textee2024}.
Through this analysis, we uncover three major categories of evaluation issues: inconsistent data processing, inconsistent task assumptions, and overly relaxed evaluation settings.
These issues largely stem from the inherent complexity of MEE, which relies on external training datasets, multi-stage pipelines, and heterogeneous modalities.
Nevertheless, such inconsistencies can lead to unfair comparisons and performance estimates that do not reflect real-world scenarios.

\subsection{Inconsistent Data Processing}
\label{subsec:inconsistent_data_processing}

Due to the absence of standardized preprocessing and the reliance on external training datasets, we observe substantial variation in data assumptions across studies.
These differences include training set construction, preprocessing, and postprocessing procedures.

\paragraph{[\textsc{P1}] Train Size Discrepancies} As described in \S\ref{subsec:preliminaries}, M2E2 relies on external datasets (e.g., ACE and SWiG) for training, which provide predefined train, development, and test splits.
While the original M2E2 benchmark uses only train splits, subsequent work often incorporates other sets as additional training data.
Consequently, models are optimized on differing numbers of samples (e.g., 75k vs. 100k images for SWiG).

\paragraph{[\textsc{P2}] Oracle Trigger Refinement} Due to distributional annotation differences between ACE and M2E2, most reported evaluation scores applies a postprocessing step that adjusts textual ED predictions using M2E2 ground truth annotations.
For example, this script removes predicted event mentions with \textit{deadly} as the trigger span. 
However, many studies do not clearly specify this experimental setting or report results exclusively with postprocessing applied.
We argue that ground truth-based postprocessing does not reflect real-world conditions.

\paragraph{[\textsc{P3}] Verb Mapping Refinement} Because the label ontologies of SWiG and M2E2 differ, the original authors provide a verb and role mapping to align SWiG annotations with the M2E2 schema.
We notice that some studies adopt refined versions of this mapping.
For example, one refined mapping aligns 73 verbs rather than the original 67 verbs to the M2E2 event types.
Such discrepancies in label alignment can lead to performance differences driven more by data engineering than by modeling improvements.

\subsection{Inconsistent Task Assumptions}
\label{subsec:inconsistent_task_assumptions}

We identify inconsistencies in task assumptions.
For instance, some studies evaluate events only present in texts and images \cite{wase2020,unicl2022,camel2023}, while others focus exclusively on multimedia events \cite{umie2024,clorae2025,ssgpf2025} (cf. \autoref{tab:methods}).
Similarly, some methods filter test data to exclude samples without events, whereas others evaluate on the full set.
Consequently, reported results are often not directly comparable.

\paragraph{[\textsc{P4}] Test Subset Selection} Recent works restrict test-time predictions to sentences or images containing at least one event.
This filtering reduces the number of sentences from 6167 to 1086 and images from 1014 to 391, while other work evaluate on the full set.
Moreover, methods focusing on the multimedia events task often only evaluate on 309 image-text pairs derived from the event coreference annotations, further reducing the test set to 192 sentences and 203 images, respectively.

\paragraph{[\textsc{P5}] MEE Task Discrepancies} 
Due to the absence of standardized evaluation scripts for multimedia events, follow up work has adopted different task definitions.
The original M2E2 benchmark considers a multimedia event correct if either the textual or visual event matches the reference and treats cross-modal coreference as a separate task.
More recent work, however, introduce a stricter setting that additionally requires correct event coreference prediction as additional attachment.
In contrast, some methods assume gold coreference links and evaluate only aligned image-text pairs.
Despite these substantial differences in task formulation, results are often compared directly.

\begin{figure}
    \centering
    \includegraphics[width=\linewidth]{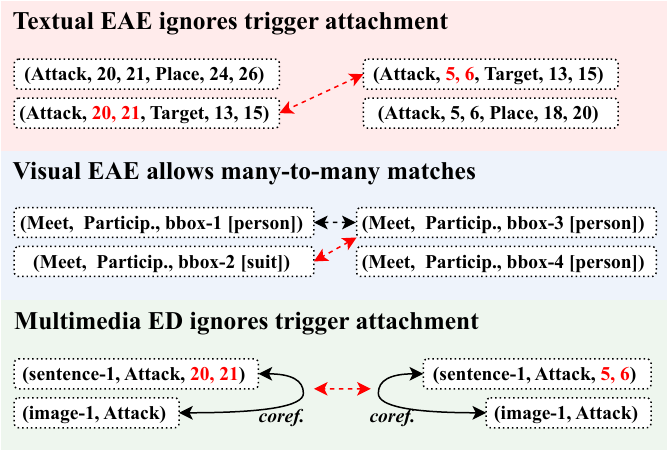}
    \caption{Illustration of relaxed evaluation settings. Red edges denote predictions that are incorrectly counted as correct, while red colored text indicates ignored trigger attachments (offsets such as \textit{20, 21}). We eliminate these issues in \textsc{StrictEval}.}
    \label{fig:relaxed_eval}
\end{figure}

\subsection{Relaxed Evaluation Settings}
\label{subsec:relaxed_evaluation_settings}

Similar to observations in TextEE \cite{textee2024}, we find that MEE evaluation metrics are often imprecise due to relaxed matching criteria or missing structural constraints. 
In \autoref{fig:relaxed_eval}, we illustrate common issues and discuss them below:

\paragraph{[\textsc{P6}] Relaxed Textual Evaluation} Some studies ignore trigger offsets during textual EAE evaluation.
When multiple events of the same type appear in a sentence, this relaxation allows arguments to be matched to any event of that type, potentially inflating reported performance such as discussed by \citet{textee2024}.

\paragraph{[\textsc{P7}] Relaxed Visual Evaluation} Most prior work employs a many-to-many matching that considers a visual argument correct if it has the correct role and its IoU exceeds a threshold.
However, this approach allows multiple predictions to match one gold argument (see \autoref{fig:duplicate_detections}), effectively rewarding recall-oriented models for visual EAE.
To address this issue, we propose a one-to-one strategy via bipartite matching, that penalizes redundant predictions (see \ref{sec:appendix:strict_veae}).

\paragraph{[\textsc{P8}] Relaxed Multimedia Evaluation} We also observe missing attachment constraints in multimedia ED evaluation.
For instance, ignoring trigger offsets allows incorrect event predictions to be counted as correct at the sentence level, rather than requiring span-level accuracy (see \autoref{fig:relaxed_eval}).
This relaxed assumption can substantially overestimate real-world performance.

\subsection{Data Leakage}
\label{subsec:other_issues}

\paragraph{[\textsc{P9}] Test Data Leakage} M2E2 forms a subset of the collected VOA samples and several works utilize this image-caption dataset during training.
We find that some of these works include images and captions from test documents in their training data.
Although no ground-truth annotations are used, exposure to test images or captions can still result in information leakage.

\subsection{\textsc{StrictEval}}
\label{subsec:strict_eval}
Our analysis demonstrates that prior MEE evaluations vary widely in data processing, task assumptions, and matching criteria. 
To systematize these discrepancies, we annotate each study with the evaluation pitfalls it exhibits (\textsc{P1}–\textsc{P9}) in \autoref{fig:framework} and cluster them into six distinct setups (see \autoref{tab:setups}).
Based on this analysis, we propose \textsc{StrictEval}, a more rigorous evaluation framework that eliminates all identified pitfalls.
\textsc{StrictEval} enforces (1) consistent data usage without oracle postprocessing or test exposure, (2) a clearly specified task definition evaluated on the full benchmark, and (3) strict matching criteria that preserve structural constraints across textual, visual, and multimedia predictions.
As shown in \autoref{tab:setups}, \textsc{StrictEval} represents the strictest setup with the goal of providing a reproducible evaluation framework designed to more faithfully reflect real-world performance.

\begin{table}
    \resizebox{\linewidth}{!}{
    \begin{tabular}{lcccccccc}
        \toprule
        \textbf{Setting} & \textsc{P1} & \textsc{P2} & \textsc{P3} & \textsc{P4} & \textsc{P5} & \textsc{P6} & \textsc{P7} & \textsc{P8} \\
        \midrule
        \textsc{StrictEval} & \xmark & \xmark & \xmark & \xmark & PC & \xmark & \xmark & \xmark \\
        \midrule
        \multicolumn{9}{l}{\textbf{\textit{Evaluation Setups}}} \\
        \textsc{Setup-1} & \xmark & \xmark & \xmark & ? & ? & \xmark & \cmark & ? \\
        \textsc{Setup-2} & \cmark & \cmark & \cmark & ? & ? & \xmark & \cmark & ? \\
        \textsc{Setup-3} & \cmark & \cmark & \cmark & \xmark & PC & \xmark & \cmark & \cmark \\
        \textsc{Setup-4}  & \cmark & \cmark & \cmark & \xmark & PC & \cmark & \cmark & \cmark \\
        \textsc{Setup-5} & \xmark & \cmark & \xmark & \cmark & PC & \xmark & \cmark & \cmark \\
        \textsc{Setup-6} & \xmark & \xmark & ? & \cmark & GC & \xmark & ? & ? \\
        \bottomrule
    \end{tabular}}
    \caption{Evaluation settings used in recent work. \textsc{P}$x$ indicates that a specific setting is used, while ? denotes that the setting is unspecified. PC evaluates with prediced coreference resolution, whereas GC uses gold coreferences.}
    \label{tab:setups}
\end{table}

\begin{table*}[ht]
    \resizebox{\linewidth}{!}{
    \begin{tabular}{llcccccccc}
        \toprule
        \multicolumn{2}{c}{} & \multicolumn{4}{c}{\textbf{ED}} & \multicolumn{4}{c}{\textbf{EAE}} \\
        \cmidrule(rl){3-6} \cmidrule(rl){7-10}
         & \textbf{Setting} & P & R & F1 & $\Delta$F1 & P & R & F1 & $\Delta$F1 \\
        \midrule
        \multirow{5}{*}{\rotatebox[origin=c]{90}{\textbf{Textual EE}}}
        & \textsc{StrictEval} & 25.8 \small{$\pm$0.6} & 75.3 \small{$\pm$1.2} & 38.4 \small{$\pm$0.6} & - & 14.2 \small{$\pm$0.3} & 48.4 \small{$\pm$0.9} & 21.9 \small{$\pm$0.3} & - \\
        & \hspace{0.5em}w/ [\textsc{P1}] train with dev & 26.5 \small{$\pm$0.4} & 73.6 \small{$\pm$0.7} & 39.0 \small{$\pm$0.5} & +0.6 & 15.0 \small{$\pm$0.4} & 47.8 \small{$\pm$0.5} & 22.8 \small{$\pm$0.5} & +0.9 \\
        & \hspace{0.5em}w/ [\textsc{P2}] trig. refinement & 41.6 \small{$\pm$1.3} & 70.4 \small{$\pm$1.2} & 52.3 \small{$\pm$1.0} & +13.9 & 20.9 \small{$\pm$0.6} & 45.9 \small{$\pm$0.9} & 28.7 \small{$\pm$0.6} & +6.8\\
        & \hspace{0.5em}w/ [\textsc{P4}] eval text subset & 57.8 \small{$\pm$0.6} & 77.5 \small{$\pm$2.0} & 66.2 \small{$\pm$0.4} & +27.8 & 24.2 \small{$\pm$0.4} & 50.1 \small{$\pm$1.3} & 32.6 \small{$\pm$0.2} & +10.7 \\
        & \hspace{0.5em}w/ [\textsc{P6}] eval EAE relaxed & - & - & - & - & 19.4 \small{$\pm$0.3} & 52.0 \small{$\pm$1.1} & 28.3 \small{$\pm$0.3} & +6.4 \\
        \midrule
        \multirow{5}{*}{\rotatebox[origin=c]{90}{\textbf{Visual EE}}}
        & \textsc{StrictEval} & 52.6 \small{$\pm$1.1} & 29.1 \small{$\pm$2.7} & 37.4 \small{$\pm$2.2} & - & 20.7 \small{$\pm$0.8} & 11.9 \small{$\pm$0.9} & 15.1 \small{$\pm$0.8} & - \\
        & \hspace{0.5em}w/ [\textsc{P1}] train with dev & 50.5 \small{$\pm$1.3} & 31.4 \small{$\pm$1.5} & 38.7 \small{$\pm$1.5} & +1.3 & 18.2 \small{$\pm$1.6} & 12.2 \small{$\pm$1.1} & 14.6 \small{$\pm$1.3} & -0.5\\
        & \hspace{0.5em}w/ [\textsc{P3}] verbs refinement & 57.6 \small{$\pm$1.5} & 36.2 \small{$\pm$1.6} & 44.4 \small{$\pm$0.8} & +7.0 & 22.9 \small{$\pm$1.7} & 15.2 \small{$\pm$0.4} & 18.2 \small{$\pm$0.6} & +3.1 \\
        & \hspace{0.5em}w/ [\textsc{P5}] eval image subset & 70.2 \small{$\pm$0.5} & 64.8 \small{$\pm$0.2} & 67.4 \small{$\pm$0.3} & +30.0 & 29.1 \small{$\pm$0.6} & 30.1 \small{$\pm$0.4} & 29.6 \small{$\pm$0.5} & +14.5 \\
        & \hspace{0.5em}w/ [\textsc{P7}] eval EAE relaxed & - & - & - & - & 21.0 \small{$\pm$0.8} & 12.2 \small{$\pm$1.0} & 15.4 \small{$\pm$0.8} & +0.3 \\
        \bottomrule
    \end{tabular}}
    \caption{Unimodal evaluation results of \textsc{Single Task} models under different setups (averaged over 3 runs). Starting from the \textsc{StrictEval} setting, each identified issue (\textsc{P}) is applied independently. The EAE relaxed settings only affect the EAE performance. $\Delta$F1 denotes the absolute difference to \textsc{StrictEval}.}
    \label{tab:systematic_results}
\end{table*}
\section{Experiments and Analysis}\label{sec:experiments}

The analysis in \S\ref{sec:method} reveals notable discrepancies and pitfalls in MEE evaluation, raising concerns about the extent to which evaluation design choices influence reported performance. 
Starting with \textsc{StrictEval}, we conduct a series of controlled experiments in which each evaluation factor is examined in isolation.

\begin{figure*}
    \centering
    \includegraphics[width=\linewidth]{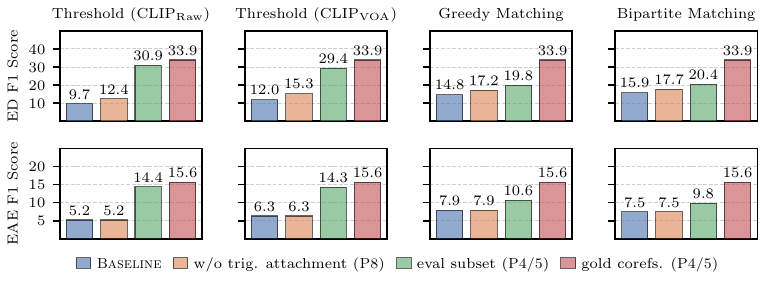}
    \caption{Multimedia ED and EAE scores using different coreference resolution techniques. Threshold Matching uses CLIP-based similarity with a threshold of 20. CLIP\textsubscript{Raw} denotes the pretrained model and CLIP\textsubscript{VOA} is further fine-tuned on the VOA image-caption dataset. Greedy Matching and Bipartite Matching follows \citet{unicl2022}.}
    \label{fig:coref_results}
\end{figure*}

\subsection{Experimental Setup}\label{subsec:experimental_setup}

\paragraph{MEE Model} We adopt \textsc{Single Task} models to avoid complex architectural choices which ensures that performance differences stem from evaluation setups rather than model capacity.
We train independent single-task models for each textual and visual subtask.
Textual and visual ED are implemented as token-level and image-level classification models.
For textual and visual EAE, we follow \cite{unicl2022,camel2023,xmtl2025} and classify ground-truth textual entities and detected visual objects (via YOLO) into argument roles.
Multimedia events are constructed through an event coreference resolution step \cite{camel2023}, described below. Further implementation details are provided in Appendix \ref{sec:appendix:mee_model} and \ref{sec:appendix:implementation}.

\paragraph{Event Coreference Resolution} Following prior work \cite{unicl2022,camel2023,mmutf2024,xmtl2025}, we perform event coreference resolution by computing CLIP-based similarity scores between image–sentence pairs \cite{clip2021}.
We note that \textsc{StrictEval} itself does not introduce a new coreference model, but instead relies on coreference predictions from existing approaches, which we compare under a unified evaluation setting.
A textual and a visual event are merged when they share the same predicted event type and their image-text similarity exceeds a threshold of 20.
We adopt this hyperparameter following \cite{camel2023,mmutf2024,xmtl2025}.
The resulting multimedia event inherits all associated arguments from both modalities.

\paragraph{Evaluation Metrics} As outlined in \S\ref{subsec:preliminaries}, we report micro-averaged P, R, and F1.
Unlike \cite{wase2020}, we follow subsequent work and require multimedia events to match event mentions and their coreference links \cite{unicl2022,camel2023,mmutf2024,xmtl2025,mgfsg2025}.
Unless otherwise specified, all experiments use our introduced evaluation framework \textsc{StrictEval.}.

\subsection{Unimodal Evaluation and Analysis}
\label{subsec:unimodal_results}

In this section, we focus on unimodal EE results.
This allows us to isolate modality-specific behavior before examining the extraction of multimedia events.
An overview of the unimodal results is presented in \autoref{tab:systematic_results} and an extended analysis in Appendix~\ref{sec:appendix:quantitative_analysis}.

\paragraph{Impact of Data Processing} Data processing choices lead to substantial performance variations.
Applying the oracle trigger refinement step [\textsc{P2}] and verb refinements [\textsc{P3}] exhibits improvements of up to +13.9 F1 for textual and +7.0 F1 for visual ED.
In contrast, incorporating the development sets of ACE and SWiG during training [\textsc{P1}] results in only marginal gains (up to +1.3 F1).
These processing decisions also yield improvements of up to +6.8 and +3.1 F1 for textual and visual EAE.
Importantly, these discrepancies extend to multimedia ED and EAE since both metrics depend on textual and visual predictions.
These findings underscore the importance of consistent data processing, as new state-of-the-art results may not solely reflect advances in modeling techniques.

\paragraph{Impact of Task Assumptions}
Inconsistent task assumptions, particularly test subset selection [\textsc{P4}], significantly boosts unimodal EE evaluation scores.
Restricting training and evaluation to texts and images that contain at least one event instance increases F1 scores of up to +27.8 and +30.0 for textual and visual ED, respectively, and improves EAE scores by up to +14.5.
These gains are primarily driven by higher precision, as no-event instances are excluded from evaluation.
Notably, this issue propagates to downstream multimedia EE and is further amplified when evaluating on smaller subsets, such as the 309 gold event coreference pairs [\textsc{P4}].
However, these results do not reflect real-world performance, where sentences and images without any targeted events are prevalent.

\paragraph{Impact of Relaxed Evaluation}
We further examine the effect of relaxed evaluation settings for textual and visual EAE, as described in \S\ref{subsec:relaxed_evaluation_settings}.
For textual EAE, ignoring the trigger span attachments [\textsc{P6}] introduces discrepancies of up to 6.4 F1.
For visual EAE, many-to-many matching [\textsc{P7}] yields only marginal improvements (+0.3 F1), however, this highly depends on the object detector and confidence threshold.
As shown in \autoref{tab:veae_results}, object detectors trained on OpenImages often predict overlapping object categories (e.g., \textit{person} and \textit{suit}), which are counted as multiple correct predictions under relaxed evaluation (e.g., +8.0 F1 for \textsc{yolo-x OI} and $\tau=0.5$).
This highlights that relaxed evaluation metrics can inflate the actual quality of EAE.

\begin{table}
    \resizebox{\linewidth}{!}{
    \begin{tabular}{lcccc}
        \toprule
        \textbf{Setting} & P & R & F1 & $\Delta$F1 \\\midrule
        \textsc{StrictEval} & 6.2 & 22.4 & 9.7 & - \\
        \hspace{0.5em} w/ train event subsets & 4.3 & 53.6 & 8.0 & -1.7 \\
        \hdashline
        \textsc{StrictEval} + eval subset & 50.4 & 22.4 & 30.9 & - \\
        \hspace{0.5em} w/ train event subsets & 53.1 & 53.6 & 53.4 & +22.2 \\
        \bottomrule
    \end{tabular}}
    \caption{Multimeda ED scores with and without training only on samples with at least one event instance.}
    \label{tab:med_results}
\end{table}

\subsection{Multimedia Evaluation and Analysis}
\label{subsec:multimedia_results}
In this section, we empirically analyze how varying task assumptions and evaluation settings affect multimedia EE results.
Real-world applications typically do not provide oracle image-text pairs or prior knowledge about whether a text or image contains a multimedia event.

\paragraph{Experimental Setup}
To study the impact of task discrepancies [\textsc{P5}] and missing trigger attachments [\textsc{P8}], we evaluate three event coreference resolution strategies: threshold-based, greedy, and bipartite matching approaches proposed in prior work \cite{unicl2022,camel2023,xmtl2025}. 
We further compare evaluations on the full dataset against subsets restricted to samples containing at least one multimedia event, and analyze the effect of applying the same subset selection during training.

\begin{table*}
    \resizebox{\linewidth}{!}{
    \begin{tabular}{llccc|ccc||ccc|ccc||cc}
        \toprule
        \multicolumn{2}{c}{} & \multicolumn{6}{c}{\textbf{\textsc{Original}}} & \multicolumn{6}{c}{\textbf{\textsc{StrictEval}}} & \multicolumn{2}{c}{\textbf{$\Delta$(\textsc{Strict} - \textsc{Orig})}} \\ 
        \multicolumn{2}{c}{} & \multicolumn{3}{c}{\textbf{ED}} & \multicolumn{3}{c}{\textbf{EAE}} & \multicolumn{3}{c}{\textbf{ED}} & \multicolumn{3}{c}{\textbf{EAE}} & \textbf{ED} & \textbf{EAE} \\ 
        \cmidrule(rl){3-5} \cmidrule(rl){6-8} \cmidrule(rl){9-11} \cmidrule(rl){12-14} \cmidrule(rl){15-15} \cmidrule(rl){16-16} 
         & \textbf{Model} & \textbf{P} & \textbf{R} & \textbf{F1} & \textbf{P} & \textbf{R} & \textbf{F1} & \textbf{P} & \textbf{R} & \textbf{F1} & \textbf{P} & \textbf{R} & \textbf{F1} & \textbf{$\Delta$F1} & \textbf{$\Delta$F1} \\
        \midrule
        \multirow{4}{*}{\rotatebox[origin=c]{90}{\textbf{Textual EE}}}
        & \textsc{Single Task} (ours) & - & - & - & - & - & - & 25.8 & 75.3 & 38.4 & 14.2 & 48.4 & 21.9 & - & - \\
        & \textsc{CAMEL} & 43.5 & 68.7 & 53.3 & 23.5 & 41.8 & 30.1 & 26.8 & 72.8 & 39.1 & 16.1 & 36.1 & 22.3 & -14.2 & ~~-7.8 \\
        & \textsc{MMUTF} & - & - & - & 31.5 & 46.8 & 37.7 & - & - & - & 15.9 & 51.7 & 24.3 & - & -13.4 \\
        & \textsc{X-MTL} & 44.2 & 65.8 & 52.9 & 29.9 & 38.9 & 33.8 & 26.6 & 70.0 & 38.5 & 20.2 & 41.7 & 27.2 & -14.4 & ~~-6.6 \\
        \midrule
        \multirow{4}{*}{\rotatebox[origin=c]{90}{\textbf{Visual EE}}}
        & \textsc{Single Task} (ours) & - & - & - & - & - & - & 52.6 & 29.1 & 37.4 & 20.7 & 11.9 & 15.1 & - & - \\
        & \textsc{CAMEL} & 66.7 & 46.5 & 54.8 & 27.7 & 19.8 & 23.1 & 55.6 & 33.0 & 41.4 & 21.7 & 11.5 & 15.0 & -13.4 & ~~-8.1 \\
        & \textsc{MMUTF} & - & - & - & 27.3 & 15.8 & 20.0 & - & - & - & 23.4 & 10.4 & 14.4 & - & ~~-5.6 \\
        & \textsc{X-MTL} & 71.7 & 69.8 & 70.7 & 31.1 & 29.6 & 30.3 & 28.2 & 69.8 & 40.2 & 14.1 & 28.8 & 18.9 & -30.5 & -11.4 \\
        \midrule
        \multirow{5}{*}{\rotatebox[origin=c]{90}{\textbf{Multi. EE}}}
        & \textsc{Single Task} (ours) & - & - & - & - & - & - & 6.2 & 22.4 & 9.7 & 3.4 & 10.8 & 5.2 & - & - \\
        & \textsc{CAMEL} & 56.2 & 42.4 & 48.3 & 29.4 & 24.0 & 26.5 & 6.9 & 27.0 & 11.0 & 3.8 & 11.0 & 5.7 & -37.3 & -20.8 \\
        & \textsc{MMUTF} & - & - & - & 37.0 & 18.7 & 24.9 & - & - & - &  4.1 & 9.1 & 5.6 & - & -19.3 \\
        & \textsc{X-MTL} & 75.5 & 57.9 & 65.6 & 36.7 & 40.6 & 38.6 & 4.9 & 43.3 & 8.8 & 2.9 & 20.0 & 5.0 & -56.8 & -33.6 \\
        & \textsc{SSGPF} & 53.7 & 70.7 & 61.0 & 15.8 & 12.6 & 14.1 & 10.3 & 30.9 & 15.4 & 2.8 & 4.2 & 3.3 & -45.6 & -10.8 \\
        \bottomrule
    \end{tabular}}
    \caption{Experimental results on the M2E2 benchmark under the original and our \textsc{StrictEval} evaluation settings. In line with \citet{mmutf2024}, the \textsc{MMUTF} model results are based on event predictions from \textsc{CAMEL}.}
    \label{tab:reproduction_results}
\end{table*}

\paragraph{Impact on Multimedia Results}
\autoref{fig:coref_results} presents the multimedia evaluation results.
Consistent with textual evaluation, ignoring trigger offsets for textual events leads to an drop of up to 3.3 F1 for ED, indicating correct sentence-level predictions with incorrect trigger spans.
Larger differences arise across task setups. Restricting evaluation to samples containing multimedia events leads to gains of up to +21.2 F1 for threshold-based methods, although such prior knowledge would not be available at deployment.
Applying the same subset selection during training further amplifies this effect and reveals opposing trends between full and subset evaluations (see \autoref{tab:med_results}).
Under realistic conditions, bipartite matching performs best, whereas threshold-based methods outperform it on the filtered subset (30.9 vs.\ 20.4 F1), attributed to increased recall.
Beyond multimedia ED, we also observe similar patterns for EAE (bottom row of \autoref{fig:coref_results}), which is connected to ED performance due to the pipeline setup.
Finally, experiments using gold image-text coreference annotations show the highest scores, highlighting that subset-based evaluation can substantially overestimate real-world performance.
\section{Consistent Evaluation}\label{sec:reproduction}

Our analysis reveals several limitations in current MEE evaluation practices, however, so far ignores the combination of hidden pitfalls and advanced modeling techniques.
Therefore, we reproduce and reevaluate recent methods \cite{camel2023,mmutf2024,xmtl2025,ssgpf2025}\footnote{Our selection focuses on methods with complete publicly available code and instructions, except for \textsc{SSGPF} as MLLM-based method.} under their original evaluation settings and compare them with our proposed framework \textsc{StrictEval}.
Notably, \textsc{SSGPF} supports only multimedia evaluation and assumes given image-text pairs.
Reproduction details are provided in Appendix~\ref{sec:appendix:reproduction_details}.

\paragraph{Reevaluation Results and Discrepancies}
The results in \autoref{tab:reproduction_results} highlight discrepancies between the \textsc{Original} and \textsc{StrictEval} setting.
First, evaluation scores change substantially in absolute performance levels.
Specifically, textual and visual ED scores differ significantly, primarily due to trigger postprocessing [\textsc{P2}], verb refinement [\textsc{P3}], and test subset selection [\textsc{P4}] (cf. \autoref{tab:systematic_results}).
This indicates that minor implementation choices can dominate reported gains.
Second, removing these pitfalls consistently leads to lower EAE performance due to the pipeline setup.
This can overestimate advances in multimedia EAE, which are often driven by increased ED performance.
Third, we observe most degradation in multimedia ED and EAE, with drops of up to 56.8 and 33.6 F1, respectively.
When neither test subset selection nor gold event coreference annotations are used, precision decreases dramatically, revealing that cross-modal event coreference resolution remains the largest challenge in a more realistic setting.
Further intuition about these low scores and performance drops is provided in Appendix~\ref{sec:appendix:quantitative_analysis}.

\paragraph{Implications for Evaluation and Future Work}
Overall, these findings underscore the need for standardized and transparent evaluation protocols.
We encourage the research community to report detailed evaluation choices, avoid reliance on gold annotations, and adopt unified evaluation pipelines to ensure fair comparison and reproducibility across MEE methods
Our analysis further indicates that progress in multimedia event extraction requires a stronger focus on accurate cross-modal event coreference and semantic alignment, which remain underexplored in recent work.
Finally, future advances would benefit from more comprehensive multimodal datasets with explicit coreference annotations, high-quality training data, and standardized splits to support robust and comparable evaluation.
\section{Conclusion}\label{sec:conclusion}

In this work, we present the first systematic analysis of evaluation pitfalls and challenges in MEE and reveal substantial gaps between reported performance and a model's actual ability to ground events across textual and visual modalities.
Our analysis of the M2E2 benchmark uncovers three major sources of discrepancies: inconsistent data processing, inconsistent task assumptions, and overly relaxed evaluation settings.
To address this, we propose the evaluation framework \textsc{StrictEval}, which enforces strict evaluation constraints for more challenging evaluation.
Controlled experiments show that minor evaluation choices can significantly affect performance, highlighting that cross-modal event coreference resolution and precise MEE remain open challenges.
\section*{Limitations}\label{sec:limitations}
In this work, we focus on analyzing evaluation pitfalls primarily based on the M2E2 benchmark, which, despite being the most widely used public benchmark for MEE, represents only a subset of possible settings.
As a result, some identified issues and recommended practices may not fully generalize to newer benchmarks, annotation schemes, or domains beyond news media.
Extending our analysis to additional datasets, modalities (e.g., videos or audio), and task formulations remains an important direction for future work.
Moreover, our empirical analysis relies on a relatively simple MEE model and a limited set of reproduced recent methods. 
While this design choice allows us to isolate the impact of evaluation decisions, it does not capture the full diversity of modeling approaches, particularly recent instruction-following MLLMs. 
Although we expect the identified evaluation pitfalls to persist across architectures, their quantitative impact may vary with different modeling paradigms.
Finally, our proposed evaluation framework adopts stricter evaluation settings and result into substantially lower absolute performance scores.
While this may hinder direct comparisons with prior work, our goal is to promote more realistic and transparent evaluation that better reflects the real-world challenges of MEE.

\section*{Ethical Considerations}\label{sec:ethics}

The results reported in this paper are intended to improve evaluation transparency in MEE and should not be interpreted as implying misconduct by prior work.
The identified evaluation pitfalls are subtle and often related to underspecified benchmark evaluation standards, making them easy to be overlooked.
Therefore, the motivation of this work is to raise awareness about these issues and to promote more reliable evaluation practices.

\bibliography{custom}

\appendix
\section{Appendix}
\label{sec:appendix}

\subsection{Articles for Systematic Analysis}
\label{sec:appendix:articles}
\autoref{tab:methods} presents a comprehensive list of studies included in our systematic analysis, along with their reported evaluation scores.
These works cover diverse modeling paradigms, ranging from early pipeline-based approaches to recent instruction-following and MLLMs.

\begin{table}[h]
    \resizebox{\linewidth}{!}{
    \begin{tabular}{lcccccc}
        \toprule
            \multicolumn{1}{c}{} & \multicolumn{2}{c}{\textbf{Textual}} & \multicolumn{2}{c}{\textbf{Visual}} & \multicolumn{2}{c}{\textbf{Multimedia}} \\ \cmidrule(rl){2-3} \cmidrule(rl){4-5} \cmidrule(rl){6-7}
            \textbf{Method} & ED & EAE & ED & EAE & ED & EAE \\\midrule
            \textsc{WASE\textsubscript{A}} (\citeyear{wase2020}) & 48.1 & 30.1 & 42.8 & 10.3 & 49.1 & 19.9 \\
            \textsc{WASE\textsubscript{O}} (\citeyear{wase2020}) & 50.6 & 26.4 & 49.9 & 11.9 & 50.8 & 19.2 \\
            \textsc{UniCL} (\citeyear{unicl2022}) & 53.7 & 30.7 & 57.6 & 15.2 & 53.4 & 23.4 \\
            \textsc{CLIP\textsubscript{Event}} (\citeyear{clipevent2022}) & - & - & 52.7 & 17.1 & - & - \\
            \textsc{CAMEL} (\citeyear{camel2023}) & 55.4 & 31.1 & 58.5 & 24.4 & 57.5 & 33.2 \\
            \textsc{UMIE} (\citeyear{umie2024}) & - & - & - & - & 62.1 & 24.5 \\
            \textsc{MGIM} (\citeyear{mgim2024}) & 55.8 & 31.2 & 58.5 & 17.8 & 55.6 & 24.6 \\
            \textsc{MMUTF} (\citeyear{mmutf2024}) & 55.5 & 38.2 & 57.0 & 20.9 & 54.6 & 27.4 \\
            \textsc{RDA} (\citeyear{rda2025}) & 56.6 & 33.4 & 60.3 & 26.1 & 58.7 & 34.6 \\
            \textsc{VEGSRF} (\citeyear{vegsrf2025}) & - & - & - & - & 53.9 & 25.3 \\
            \textsc{AFSSAG} (\citeyear{afssag2025}) & - & - & - & - & 54.4 & 26.5 \\
            \textsc{C-LoRAE} (\citeyear{clorae2025}) & - & - & - & - & 63.5 & 32.6 \\
            \textsc{X-MTL} (\citeyear{xmtl2025}) & 56.6 & 36.0 & 71.7 & 32.2 & 66.2 & 41.4 \\
            \textsc{MGFSG} (\citeyear{mgfsg2025}) & 58.5 & 31.9 & 61.8 & 20.3 & 57.9 & 27.4 \\
            \textsc{MLLM} (\citeyear{mllm-ke2025}) & 76.0 & 39.3 & 65.3 & 27.3 & 87.4 & 44.3 \\
            \textsc{SSGPF} (\citeyear{ssgpf2025}) & - & - & - & - & 65.7 & 36.0 \\
        \bottomrule
    \end{tabular}}
    \caption{Reported F1 scores of methods on M2E2.}
    \label{tab:methods}
\end{table}

\subsection{One-to-One Matching Strategy}
\label{sec:appendix:strict_veae}

As described in \S\ref{subsec:relaxed_evaluation_settings}, visual arguments are initially evaluated using a many-to-many matching strategy.
In this setting, multiple predicted arguments may be matched to the same ground-truth argument (many-to-one) and a single predicted argument may match multiple ground-truth arguments (one-to-many).
While this approach captures all overlapping predictions, it can inflate evaluation scores and does not enforce a strict mapping between predicted and gold arguments.

\paragraph{One-to-One Matching}
To address this limitation, we adopt a one-to-one matching strategy based on the Hungarian algorithm \cite{hungarian1955}.
Predicted and ground-truth arguments are modeled as nodes in a bipartite graph and a globally optimal matching is computed between the two sets.
This formulation guarantees that each predicted argument is matched to at most one ground-truth argument, yielding a stricter and more interpretable evaluation.

\begin{figure*}
    \centering
    \includegraphics[width=\linewidth]{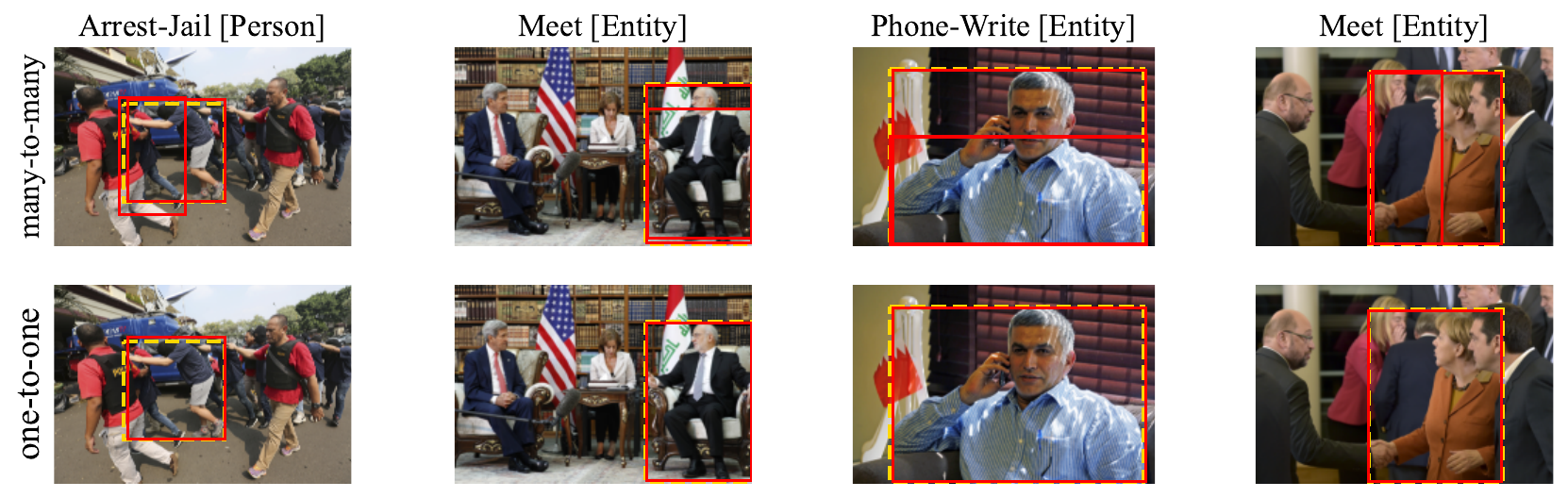}
    \caption{Qualitative error analysis of visual EAE. Gold rectangles denote ground-truth argument roles while red rectangles indicate correctly counted predictions. The top row illustrates a common failure case in which overlapping objects (often other persons, suits, or shirts) are incorrectly counted as matches. This inflates performance metrics such as recall. In contrast, the bottom row shows our proposed one-to-one version which alleviates these error cases.}
    \label{fig:duplicate_detections}
\end{figure*}

\paragraph{Empirical Impact Analysis}
In \autoref{tab:veae_results}, we show the impact for object detectors YOLOv8 \cite{yolo2016,yolo_v8_2024} and Faster R-CNN \cite{fasterrcnn2014}, trained on COCO \cite{coco2014} and OpenImages \cite{openimages2020}, respectively.
Models trained on COCO show robust performance under the stricter matching strategy, whereas results on OpenImages exhibit substantial differences.
For example, we observe an F1 increase of +8.0 for \textsc{YOLO-X OI} at $\tau=0.5$.
This behavior can be attributed to overlapping object categories in OpenImages (e.g., \textit{person} and \textit{suit}) which are otherwise counted as multiple correct predictions under many-to-many matching (see \autoref{fig:duplicate_detections}).

\begin{table}
    \resizebox{\linewidth}{!}{
    \begin{tabular}{lcccccccc}
        \toprule
        \multicolumn{3}{c}{} & \multicolumn{3}{c}{\textbf{many-to-many}} & \multicolumn{3}{c}{\textbf{one-to-one}} \\ \cmidrule(rl){4-6} \cmidrule(rl){7-9}
        \textbf{Detector} & \textbf{$\tau$} & \textbf{\#Pred} & P & R & F1 & P & R & F1 \\
        \midrule
        \textsc{yolo-x CC} & 0.8 & 1617 & 32.5 & 36.8 & 34.5 & 31.5 & 35.7 & 33.5 \\
        \textsc{yolo-x CC} & 0.5 & 3162 & 20.9 & 46.3 & 28.8 & 20.1 & 44.4 & 27.6 \\
        \textsc{yolo-x CC} & 0.1 &  5067 & 14.6 & 51.9 & 22.8 & 13.4 & 47.7 & 21.0 \\
        \hdashline
        \textsc{yolo-x OI} & 0.8 & 267 & 51.7 & 9.7 & 16.3 & 45.7 & 8.5 & 14.4 \\
        \textsc{yolo-x OI} & 0.5 & 2349 & 26.5 & 43.6 & 33.0 & 20.1 & 33.0 & 25.0 \\
        \textsc{yolo-x OI} & 0.1 & 5917 & 17.8 & 73.7 & 28.7 & 11.1 & 45.8 & 17.8 \\
        \hdashline
        \textsc{fr-cnn OI} & 0.8 & 191 & 36.7 & 4.9 & 8.6 & 35.6 & 4.7 & 8.4\\
        \textsc{fr-cnn OI} & 0.5 & 1037 & 37.1 & 26.9 & 31.2 & 30.3 & 21.9 & 25.5 \\
        \textsc{fr-cnn OI} & 0.1 & 4086 & 28.8 & 82.4 & 42.7 & 12.7 & 36.3 & 18.8 \\
        \bottomrule
    \end{tabular}}
    \caption{Visual EAE scores using ground truth events with the many-to-many (original) and one-to-one (ours) evaluation. Models labeled \textsc{CC} and \textsc{OI} correspond to object detectors trained on COCO (80 classes) and OpenImages (600 classes), respectively. The parameter $\tau$ denotes the chosen minimum confidence threshold for each object.}
    \label{tab:veae_results}
\end{table}

\subsection{Proposed \textsc{Single Task} Models}
\label{sec:appendix:mee_model}

In this section, we describe the \textsc{Single Task} models used in the experiments reported in \S\ref{sec:experiments}.
Each subtask is modeled independently, including textual ED / EAE, visual ED / EAE, and Event Coreference Resolution.

\paragraph{Textual Event Extraction} 
For textual ED and EAE, we employ BERT \cite{bert2019} as the text encoder.
Given an input sentence $s$, BERT produces contextualized subtoken representations, which are mean-pooled to obtain token-level embeddings.
Textual ED is formulated as a sequence labeling task where each token is classified into an event type using a linear classifier.
For textual EAE, following \citet{wase2020}, we assume gold entity mentions and perform role classification by mean-pooling the subtoken representations of each entity mention.
Textual ED and EAE are trained as separate models using cross-entropy loss.

\paragraph{Visual Event Extraction}
For visual ED and EAE, we adopt the CLIP \cite{clip2021} vision encoder.
Given an input image $i$, CLIP produces global and patch-level representations.
Visual ED is treated as image-level classification by feeding the [CLS] token representation into a linear classifier.
For visual EAE, we first detect objects using an offline detector \cite{xmtl2025}.
Patch representations corresponding to each object are mean-pooled to form object embeddings, which are then classified into argument roles using a linear layer.
Analogous to the textual tasks, Visual ED and EAE are trained independently with cross-entropy loss.

\paragraph{Multimedia Event Extraction} 
As described in \S\ref{subsec:experimental_setup}, event coreference resolution between textual and visual events is performed using CLIP-based similarity scores.
Following prior work \cite{camel2023,mmutf2024,xmtl2025}, we construct a multimedia event when a text–image pair shares the same predicted event type and their similarity score exceeds a threshold of 20.
The resulting multimedia event aggregates all associated textual and visual arguments.
In addition to this heuristic approach, we also report results obtained using greedy and bipartite matching strategies \cite{unicl2022}.

\paragraph{Training}
We train all models using a learning rate of $1\times10^{-5}$ for encoder parameters and $1\times10^{-4}$ for classifier parameters.
Textual models are trained with a batch size of 16 for 20 epochs, while visual models use a batch size of 64 and are trained for 10 epochs.
Model performance is evaluated at the end of each epoch on the ACE development set for textual tasks and the SWiG development set for visual tasks.
We select the checkpoint with the best development set performance for final evaluation.

\subsection{Implementation Details}\label{sec:appendix:implementation} 

All models are implemented using the \textit{Transformers} \citep{transformers2020} (v4.55.0) library in conjunction with \textit{PyTorch} (v2.8.0).
Unless otherwise specified, we use \textit{bert-base-uncased} \cite{bert2019} with 222M parameters as text encoder and \textit{clip-vit-base-patch16} \cite{clip2021} with 85M parameters as vision encoder.
Object detections are obtained using YOLOv8 \cite{yolo_v8_2024}\footnote{\url{https://docs.ultralytics.com/models/yolov8}} trained on COCO and detections with confidence scores below 0.8 are discarded.
All experiments are conducted on NVIDIA A100 GPUs within a single compute node running CUDA 12.3.
We run each experiment with three random seeds and report the average performance across runs.

\subsection{Reproduction Details}
\label{sec:appendix:reproduction_details}

In this section, we provide detailed reproduction information for the models discussed in \S\ref{sec:reproduction}, along with potential explanations for any observed differences in performance scores.

\paragraph{\textsc{CAMEL}} 
We use the official released code\footnote{\url{https://github.com/ZILIN003/CAMEL}} provided by \citet{camel2023}. 
Our reproduced F1 scores largely align with the reported results, except for multimedia ED (48.3 vs. 57.5 F1) and EAE (26.5 vs 33.2 F1). 
We attribute these discrepancies primarily to the absence of balanced visual ED training in our reproduction that reduced recall in favor of precision.

\paragraph{\textsc{MMUTF}}
We use the official released code\footnote{\url{https://github.com/seebergerph/MMUTF}} provided by \citet{mmutf2024}. 
Following the original paper, which mainly focuses on EAE, we use ED predictions from \textsc{CAMEL}. 
Our reproduced scores closely match the reported results with minor decreases in multimedia ED and EAE, which we also attribute to the lower recall of \textsc{CAMEL} predictions.

\paragraph{\textsc{X-MTL}} 
We use the official released code\footnote{\url{https://github.com/aoine-dev/X-MTL}} provided by \citet{xmtl2025}. 
Our reproduced results show only minor deviations from the originally reported results (e.g., 56.6 vs. 52.9 F1 for textual ED). 
We believe these differences are related to inconsistencies in the pseudo-labeled VOA image-caption dataset, caused by broken links during dataset construction.

\paragraph{\textsc{SSGPF}} We re-implement \textsc{SSGPF} using \textit{LLaVA-v1.5-7B} as described in the paper\footnote{\url{https://github.com/MartinYuanNJU/SSGPF}} \cite{ssgpf2025}.
The model assumes aligned image-sentence pairs and requires manually written event and role descriptions.
We obtain comparable performance on multimedia ED (65.7 vs. 61.0 F1) but observe a substantial drop on EAE (36.0 vs. 14.1 F1), which we attribute to implementation differences in EAE evaluation, visual grounding (SEEM), and role descriptions.
For \textsc{StrictEval}, we use the proposed fine-tuned cross-modal retrieval model to construct image-sentence pairs.

\subsection{Analysis of \textsc{StrictEval}}
\label{sec:appendix:quantitative_analysis}
We base our experiments on the human-annotated \textsc{M2E2} dataset (Inter-Annotator
Agreement of 81.2\%) \cite{wase2020} and remove relaxed assumptions used in prior work to ensure more consistent comparison.
The resulting substantially lower scores are primarily due to stricter evaluation protocols that expose false positives otherwise ignored, which we detail next. Using our proposed \textsc{Single Task} models, we provide further intuition for these effects.

\paragraph{Textual Evaluation}
In textual evaluation, oracle trigger refinement\footnote{\url{https://github.com/jianliu-ml/Multimedia-EE/blob/main/code/textualEE/refine_result.py}} and test subset selection remove a large number of incorrect predictions and negative samples.
For example, the \textsc{Single Task} model predicts 3,969 text events, of which refinement removes 1,636 false positive events and 1,885 arguments prior to evaluation.
Furthermore, our manual analysis suggests annotation discrepancies between \textsc{ACE} and \textsc{M2E2}, potentially contributing to the large number of false positive events.

\paragraph{Visual Evaluation}
Similarly, test subset selection excludes 623 images without annotated events, restricting evaluation to only 391 positive samples. 
As a result, false positives from the excluded images are not counted. 
For instance, the \textsc{Single Task} model predicts 88 false positive visual events and 341 arguments on these omitted images. 
Evaluating only on positive samples therefore inflates precision by ignoring predictions on negative images.

\paragraph{Multimedia Evaluation}
For multimedia evaluation, recent work often assumes access to ground truth image–sentence pairs or applies post-hoc filtering using gold sentence and image IDs (e.g., via coreference annotations). 
This substantially reduces the number of evaluated pairs and removes negative candidates. 
In contrast, \textsc{StrictEval} constructs and evaluates over all possible image–sentence pairs. 
Consequently, models must handle a much larger and noisier candidate space, leading to a significant increase in false positives.
For our baseline models, post-hoc filtering excludes 1,118 false positive multimedia events, resulting into a substantial drop in precision and F1 scores.

\subsection{Additional Experimental Results}
\label{sec:appendix:full_mee_results}

This section supplements the multimedia results presented in \S\ref{subsec:multimedia_results} by providing EAE scores along with precision and recall. 
We report evaluations for different coreference resolution strategies: threshold-based (\autoref{tab:mee_results_threshold_raw} and \autoref{tab:mee_results_threshold_voa}), greedy matching (\autoref{tab:mee_results_greedy}), and bipartite matching (\autoref{tab:mee_results_bipartite}).

\begin{table*}
    \resizebox{\linewidth}{!}{
    \begin{tabular}{llcccccccc}
        \toprule
        \multicolumn{2}{c}{} & \multicolumn{4}{c}{\textbf{ED}} & \multicolumn{4}{c}{\textbf{EAE}} \\
        \cmidrule(rl){3-6} \cmidrule(rl){7-10}
         & \textbf{Setting} & P & R & F1 & $\Delta$F1 & P & R & F1 & $\Delta$F1 \\
        \midrule
        \multirow{4}{*}{\rotatebox[origin=c]{90}{\textbf{Multi. EE}}}
        & \textsc{StrictEval} & 6.2 \small{$\pm$0.9} & 22.4 \small{$\pm$3.1} & 9.7 \small{$\pm$1.4} & - & 3.4 \small{$\pm$0.4} & 10.8 \small{$\pm$1.1} & 5.2 \small{$\pm$0.5} & - \\
        & \hspace{0.5em}w/ [\textsc{P8}] eval MED relaxed & 8.4 \small{$\pm$1.2} & 23.7 \small{$\pm$3.3} & 12.4 \small{$\pm$1.6} & +2.7 & 3.4 \small{$\pm$0.4} & 10.8 \small{$\pm$1.1} & 5.2 \small{$\pm$0.5} & - \\
        & \hspace{0.5em}w/ [\textsc{P4/5}] eval text/image subset & 50.4 \small{$\pm$3.7} & 22.4 \small{$\pm$3.1} & 30.9 \small{$\pm$3.3} & +21.2  & 22.9 \small{$\pm$0.9} & 10.8 \small{$\pm$1.1} & 14.4 \small{$\pm$1.0} & +9.2 \\
        & \hspace{0.5em}w/ [\textsc{P4/5}] gold coreferences & 66.5 \small{$\pm$2.5} & 22.8 \small{$\pm$3.2} & 33.9 \small{$\pm$3.8} & +24.2 & 27.5 \small{$\pm$0.7} & 11.3 \small{$\pm$1.3} & 15.6 \small{$\pm$1.3} & +10.4 \\
        \bottomrule
    \end{tabular}}
    \caption{Evaluation results of MEE model with threshold-based coreference resolution (CLIP\textsubscript{Raw}).}
    \label{tab:mee_results_threshold_raw}
\end{table*}

\begin{table*}
    \resizebox{\linewidth}{!}{
    \begin{tabular}{llcccccccc}
        \toprule
        \multicolumn{2}{c}{} & \multicolumn{4}{c}{\textbf{ED}} & \multicolumn{4}{c}{\textbf{EAE}} \\
        \cmidrule(rl){3-6} \cmidrule(rl){7-10}
         & \textbf{Setting} & P & R & F1 & $\Delta$F1 & P & R & F1 & $\Delta$F1 \\
        \midrule
        \multirow{4}{*}{\rotatebox[origin=c]{90}{\textbf{Multi. EE}}}
        & \textsc{StrictEval} & 8.6 \small{$\pm$1.1} & 20.4 \small{$\pm$2.8} & 12.0 \small{$\pm$1.5} & - & 4.6 \small{$\pm$0.5} & 10.2 \small{$\pm$1.0} & 6.3 \small{$\pm$0.5} & - \\
        & \hspace{0.5em}w/ [\textsc{P8}] eval MED relaxed & 11.9 \small{$\pm$1.4} & 21.7 \small{$\pm$3.1} & 15.3 \small{$\pm$1.7} & +3.3 & 4.6 \small{$\pm$0.5} & 10.2 \small{$\pm$1.0} & 6.3 \small{$\pm$0.5} & - \\
        & \hspace{0.5em}w/ [\textsc{P4/5}] eval text/image subset & 52.9 \small{$\pm$2.6} & 20.4 \small{$\pm$2.8} & 29.4 \small{$\pm$3.3} & +17.4 & 24.0 \small{$\pm$0.1} & 10.2 \small{$\pm$1.0} & 14.3 \small{$\pm$1.0} & +8.0 \\
        & \hspace{0.5em}w/ [\textsc{P4/5}] gold coreferences & 66.5 \small{$\pm$2.5} & 22.8 \small{$\pm$3.2} & 33.9 \small{$\pm$3.8} & +21.9 & 27.5 \small{$\pm$0.7} & 11.3 \small{$\pm$1.3} & 15.6 \small{$\pm$1.3} & +9.3 \\
        \bottomrule
    \end{tabular}}
    \caption{Evaluation results of MEE model with threshold-based coreference resolution (CLIP\textsubscript{VOA}).}
    \label{tab:mee_results_threshold_voa}
\end{table*}

\begin{table*}
    \resizebox{\linewidth}{!}{
    \begin{tabular}{llcccccccc}
        \toprule
        \multicolumn{2}{c}{} & \multicolumn{4}{c}{\textbf{ED}} & \multicolumn{4}{c}{\textbf{EAE}} \\
        \cmidrule(rl){3-6} \cmidrule(rl){7-10}
         & \textbf{Setting} & P & R & F1 & $\Delta$F1 & P & R & F1 & $\Delta$F1 \\
        \midrule
        \multirow{4}{*}{\rotatebox[origin=c]{90}{\textbf{Multi. EE}}}
        & \textsc{StrictEval} & 20.0 \small{$\pm$0.8} & 11.8 \small{$\pm$1.1} & 14.8 \small{$\pm$0.9} & - & 10.0 \small{$\pm$0.8} & 6.6 \small{$\pm$0.7} & 7.9 \small{$\pm$0.7} & - \\
        & \hspace{0.5em}w/ [\textsc{P8}] eval MED relaxed & 26.8 \small{$\pm$1.0} & 12.7 \small{$\pm$1.3} & 17.2 \small{$\pm$1.2} & +2.4 & 10.0 \small{$\pm$0.8} & 6.6 \small{$\pm$0.7} & 7.9 \small{$\pm$0.7} & - \\
        & \hspace{0.5em}w/ [\textsc{P4/5}] eval text/image subset & 60.5 \small{$\pm$3.1} & 11.8 \small{$\pm$1.1} & 19.8 \small{$\pm$1.7} & +5.0 & 27.1 \small{$\pm$1.1} & 6.6 \small{$\pm$0.7} & 10.6 \small{$\pm$1.0} & +2.7 \\
        & \hspace{0.5em}w/ [\textsc{P4/5}] gold coreferences & 66.5 \small{$\pm$2.5} & 22.8 \small{$\pm$3.2} & 33.9 \small{$\pm$3.8} & +14.1 & 27.5 \small{$\pm$0.7} & 11.3 \small{$\pm$1.3} & 15.6 \small{$\pm$1.3} & +7.7 \\
        \bottomrule
    \end{tabular}}
    \caption{Evaluation results of MEE model with greedy matching coreference resolution.}
    \label{tab:mee_results_greedy}
\end{table*}

\begin{table*}
    \resizebox{\linewidth}{!}{
    \begin{tabular}{llcccccccc}
        \toprule
        \multicolumn{2}{c}{} & \multicolumn{4}{c}{\textbf{ED}} & \multicolumn{4}{c}{\textbf{EAE}} \\
        \cmidrule(rl){3-6} \cmidrule(rl){7-10}
         & \textbf{Setting} & P & R & F1 & $\Delta$F1 & P & R & F1 & $\Delta$F1 \\
        \midrule
        \multirow{4}{*}{\rotatebox[origin=c]{90}{\textbf{Multi. EE}}}
        & \textsc{StrictEval} & 23.3 \small{$\pm$1.7} & 12.1 \small{$\pm$1.3} & 15.9 \small{$\pm$1.5} & - & 10.1 \small{$\pm$0.8} & 6.0 \small{$\pm$0.8} & 7.5 \small{$\pm$0.8} & - \\
        & \hspace{0.5em}w/ [\textsc{P8}] eval MED relaxed & 29.0 \small{$\pm$1.2} & 12.8 \small{$\pm$1.5} & 17.7 \small{$\pm$1.6} & +1.8 & 10.1 \small{$\pm$0.8} & 6.0 \small{$\pm$0.8} & 7.5 \small{$\pm$0.8} & - \\
        & \hspace{0.5em}w/ [\textsc{P4/5}] eval text/image subset & 67.5 \small{$\pm$3.6} & 12.1 \small{$\pm$1.3} & 20.4 \small{$\pm$2.1} & +4.5 & 26.4 \small{$\pm$1.4} & 6.0 \small{$\pm$0.8} & 9.8 \small{$\pm$1.2} & +2.3 \\
        & \hspace{0.5em}w/ [\textsc{P4/5}] gold coreferences & 66.5 \small{$\pm$2.5} & 22.8 \small{$\pm$3.2} & 33.9 \small{$\pm$3.8} & +18.0 & 27.5 \small{$\pm$0.7} & 11.3 \small{$\pm$1.3} & 15.6 \small{$\pm$1.3} & +8.1 \\
        \bottomrule
    \end{tabular}}
    \caption{Evaluation results of MEE model with bipartite matching coreference resolution.}
    \label{tab:mee_results_bipartite}
\end{table*}

\end{document}